\documentclass[conference]{IEEEtran}
\IEEEoverridecommandlockouts
\usepackage{cite}
\usepackage{amsmath, amssymb, amsfonts}
\usepackage{algorithmic}
\usepackage{graphicx}
\usepackage{textcomp}
\usepackage{xcolor}
\usepackage{pgfplots}
\usepackage{tikz}
\usetikzlibrary{arrows.meta, positioning, shapes.geometric}
\def\BibTeX{{\rm B\kern-.05em{\sc i\kern-.025em b}\kern-.08em T\kern-.1667em\lower.7ex\hbox{E}\kern-.125emX}}
\newcommand{\Chronos}{\textsc{Chronos-2}}
\begin{document}
    \csname title\endcsname{Time-Series Foundation Model Embeddings for Remaining Useful Life Estimation}

    \author{ \IEEEauthorblockN{ Amir El-Ghoussani\IEEEauthorrefmark{1}\quad Michele De Vita\IEEEauthorrefmark{1}\quad Ronald Naumann\IEEEauthorrefmark{2}\quad Vasileios Belagiannis\IEEEauthorrefmark{1} }
    \IEEEauthorblockA{\IEEEauthorrefmark{1} \textit{Friedrich-Alexander-Universit\"at Erlangen--N\"urnberg (FAU), Germany}\\ \textit{\{amir.el-ghoussani, michele.de.vita, vasileios.belagiannis\}@fau.de} }
    \IEEEauthorblockA{\IEEEauthorrefmark{2} \textit{Nokia Solutions and Networks GmbH \& Co. KG, Germany}\\ \textit{ronald.naumann@nokia.com} }
    }

    \maketitle

    \begin{abstract}
        Remaining Useful Life (RUL) prediction is essential for industrial predictive
        maintenance, yet many learning-based approaches rely on extensive
        feature engineering or large labeled datasets to train task-specific
        sequence models. In this work, we introduce a lightweight learning
        approach, in which we leverage a frozen pretrained time-series foundation
        model (TSFM) and combine it with a small regression head for RUL estimation
        from multivariate sensor streams. More specifically, we use Chronos-2 as
        a frozen backbone to extract context window features and train a lightweight
        regression neural network for RUL prediction. Experiments on real-world
        industrial sensor data from two device types show that \Chronos\
        features consistently improve over recurrent, convolutional, Transformer-based,
        and gradient-boosting baselines under the same preprocessing and evaluation
        protocol. We further analyze the impact of context length and find that
        performance improves significantly with longer histories, indicating that
        TSFM representation offer a practical and data-efficient alternative for
        RUL estimation in industrial settings.
    \end{abstract}


    \section{Introduction}
    Predictive maintenance aims to forecast equipment failures, so that interventions
    can be planned in advance to minimise unexpected downtime and associated costs.
    A key metric in this context is the Remaining Useful Life (RUL) that is
    defined as the estimated time until an asset's failure based on its current
    condition. Predicting device RUL based on multivariate sensor streams accurately
    is vital for effective maintenance, and is therefore a major focus in the
    prognostics field~\cite{jardine2006review,si2011rulreview}.

    Traditional data-driven RUL estimation usually relies on either handcrafted
    features or end-to-end sequence models, which have been trained using multivariate
    historical sensor measurements~\cite{saxena2008cmapss}. For one-dimensional sensor signals, encoder-decoder denoising with adversarial latent alignment have been used to learn noise-robust representations~\cite{casas_adversarial}. While deeper
    sequence models such as long short-term memory (LSTM) networks~\cite{hochreiter1997lstm}
    and gated recurrent units (GRU)~\cite{cho2014gru} can capture temporal
    patterns, they often require substantial labeled data and careful tuning for
    different sensor types. To address these limitations, temporal convolutional
    neural networks (CNNs)~\cite{li2018cnn_rul} and Transformer-based~\cite{chronos}
    models have been proposed for representation learning and long-range dependencies. 
    Transformer-style architectures with hierarchical or multiscale attention are
    increasingly used to model sensor interactions over time in multivariate signal
    streams~\cite{fan2024star}. To relax the labeling requirements, contrastive and
    self-supervised representation learning has also been used to pretrain on
    unlabeled sensor data before supervised RUL regression~\cite{fu2024dualmixer}. Related label-efficient temporal monitoring problems have also been studied in unsupervised anomaly detection, where models learn normal multi-agent trajectory behavior from unlabeled data and are evaluated on annotated abnormal scenarios~\cite{wiederer_benchmark}.
    Robustness to distribution shift under varying operating conditions is often
    tackled via adversarial domain adaptation~\cite{du2024madan}. In parallel,
    physics-informed hybrids introduce degradation structure to improve
    extrapolation and interpretability~\cite{jiang2024stahpinn}. Generative diffusion
    approaches are explored for data augmentation and uncertainty modeling, both
    via diffusion models for generating realistic degradation trajectories
    \cite{wang2024diffrul} and stochastic diffusion-process models that predict
    RUL distributions \cite{wen2025diffusion}. However, these approaches require
    training on task-specific data, which can be resource-intensive and may not generalize
    well across different datasets. In contrast, we find that time-series foundation
    models (TSFMs) provide a way to obtain temporally rich representations without
    the need for extensive training.

    \begin{figure}[t]
        \includegraphics[width=.5\linewidth]{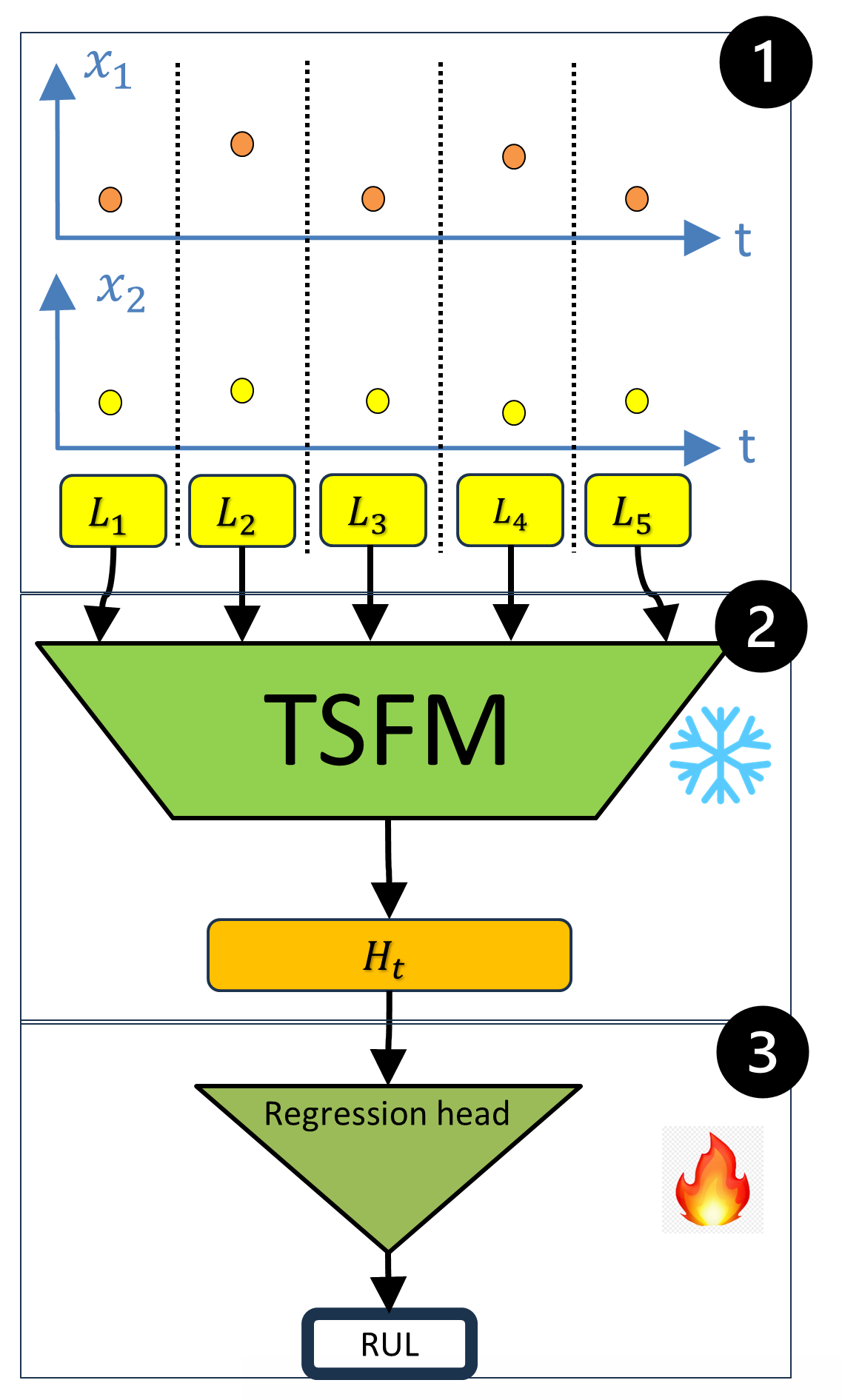}
        \begin{tikzpicture}[remember picture, overlay]
            \node[anchor=north, inner sep=0] (image) at (0.5,0) { };
            \node[anchor=north, font=\small]
                at
                (2,7)
                { We subdivide the multivariate sensor
                };
            \node[anchor=north, font=\small]
                at
                (2,6.5)
                {
                stream into context windows of
                };
            \node[anchor=north, font=\small]
                at
                (2,6)
                {

                length $L$. };
            \node[anchor=north, font=\small]
                at
                (2,4.5)
                { We feed each window into a frozen

                };
            \node[anchor=north, font=\small]
                at
                (2,4)
                {
                pretrained TSFM to extract temporal
                };
            \node[anchor=north, font=\small]
                at
                (2,3.5)
                {

                context features. };
            \node[anchor=north, font=\small]
                at
                (2,2)
                { We train a lightweight regression

                };
            \node[anchor=north, font=\small]
                at
                (2,1.5)
                {
                head with embeddings $\mathbf{H}_{t}$
                };
            \node[anchor=north, font=\small]
                at
                (2,1)
                {

                features to predict RUL. };
        \end{tikzpicture}
        \caption{Overview of our lightweight approach to RUL estimation on
        industrial data.}
        \label{fig:teaser}
    \end{figure}
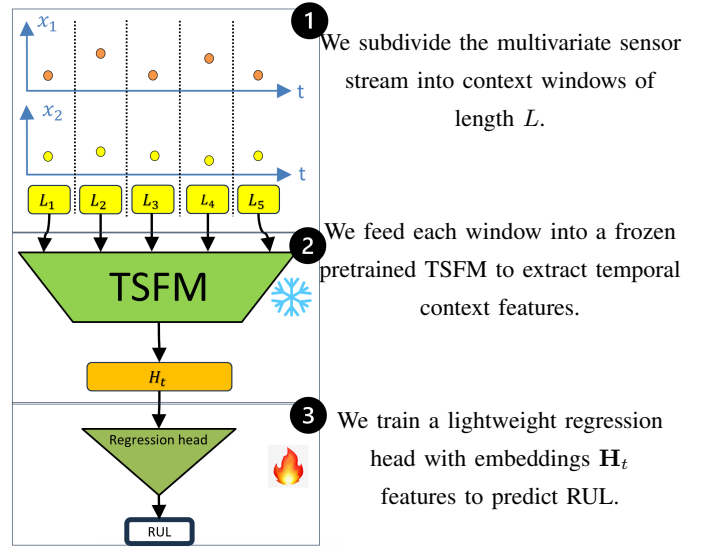

    TSFMs are predominantly trained and used for time-series forecasting tasks. Recent
    approaches such as Chronos~\cite{chronos} and its successor \Chronos~\cite{chronos2}
    are probabilistic transformer-based TSFMs. Both versions are built on a
    decoder-only transformer architecture that tokenizes input time series into
    discrete bins. \Chronos \ enables efficient in-context learning without requiring
    fine-tuning. Kairos~\cite{feng2025kairos} uses masked token modeling to
    learn representations for multivariate time series. VisionTS++~\cite{shen2025visionts++}
    applies vision transformer architectures to time-series patches for improved
    long-range modeling. TimesFM~\cite{das2024decoder} leverages large-scale
    pretraining with a masked forecasting objective. PatchTST~\cite{nie2022time}
    divides time series into patches and processes them with transformers for efficient
    sequence modeling. MiroAI~\cite{woo2024unified}, is based on masked autoencoding
    and large-scale pretraining.

    In this work, we introduce an approach leveraging pretrained TSFM
    representations for RUL estimation from large-scale sensor streams. We extend
    TSFMs beyond their traditional forecasting use case by treating RUL prediction
    as a supervised regression problem. We adapt {\Chronos} by loading a look-back
    window of past sensor measurements directly into the model's context. From this
    context, we extract backbone representations and train a lightweight
    regression head on top, keeping the adaptation simple and efficient.
    Experiments on a large-scale industrial sensor dataset show that
    conditioning \Chronos\ on sensor-history context yields substantial gains, outperforming
    standard baselines---including non-sequential regressors---by up to $5\times$.
    We further find that context length is critical: expanding the context window
    to 80 steps dramatically improves performance by reducing MAE by $2\times$ compared
    to a short {5-step} context.

    \section{Method}

    To predict the RUL, we process the complete history of multivariate sensor data.
    Let $\mathbf{X}\in \mathbb{R}^{T \times D}$ represent the entire sequence of
    sensor measurements:
    \begin{equation}
        \mathbf{X}= [\mathbf{x}_{1}, \ldots, \mathbf{x}_{T}],
    \end{equation}
    where each $\mathbf{x}_{t}$ is a $D$-dimensional vector of sensor readings at
    time $t$, and $T$ is the total length of the time series.

    The goal is to estimate the RUL, denoted as $y_{t}$, for each time step $t$.
    We approach this by mapping the input sequence to a sequence of RUL estimates
    via a combination of 2 models, consisting of a frozen pre-trained backbone
    $\Phi(\cdot)$ and a trainable regression head $g_{\phi}(\cdot)$:
    \begin{equation}
        \hat{y}_{t}= g_{\phi}(\Phi(\mathbf{X})_{t}).
    \end{equation}
    The following subsections detail the data processing pipeline, the specific
    architecture of the backbone and head, and our training protocol.

    \subsection{Data Preprocessing and Label Generation}
    \label{sec:method_data}

    Raw sensor measurements are recorded at irregular timestamps due to network
    delays and asynchronous hardware. However, most pretrained TSFMs, including \textsc{Chronos},
    require regularly sampled data. For this reason, we linearly interpolate each
    sensor record to a uniform time grid with step size $\Delta t$.

    Given irregular timestamps $\tau_{k}$ and $\tau_{k+1}$, we linearly
    interpolate the value $x_{t}$ at a regular time $t$ (where $\tau_{k}\le t \le
    \tau_{k+1}$) as:
    \begin{equation}
        x_{t}= x_{\tau_k}+ (x_{\tau_{k+1}}- x_{\tau_k}) \frac{t - \tau_{k}}{\tau_{k+1}-
        \tau_{k}}.
    \end{equation}
    This yields a multivariate sequence with regular intervals $\{\mathbf{x}_{t}\}
    _{t=1}^{T}$. To prevent artifacts from long outages, we discard intervals where
    the gap between consecutive raw samples is higher than a threshold $\Delta t_{\max}$
    ($\tau_{k+1}- \tau_{k}> \Delta t_{\max}$) and only interpolate within valid ranges.

    Following resampling, we filter out faulty measurements containing \texttt{NaN}
    values. We apply a global outlier clipping strategy, clamping values to the 1st
    and 99th percentiles, and normalize sensor channels to zero mean and unit
    variance. All statistics are computed exclusively on the training split to prevent
    data leakage.

    \paragraph{RUL Label Construction}
    In addition to sensor measurements, we utilize maintenance logs containing the
    set of failure timestamps $\mathcal{T}_{\mathrm{rep}}$. For a given
    timestamp $t$, the time of the next failure is defined as:
    \begin{equation}
        t_{\mathrm{fail}}=\min\{t' \in \mathcal{T}_{\mathrm{rep}}: t' > t\}.
    \end{equation}
    The RUL label is computed as the time remaining until this event:
    \begin{equation}
        y_{t}= t_{\mathrm{fail}}- t.
    \end{equation}
    We express $y_{t}$ in days and cap such that $y_{t}\leftarrow \min(y_{t}, y_{\max}
    )$, with $y_{\max}=1000$ days. Timestamps with no subsequent repair events are
    excluded from the supervised training set.

    \subsection{Model Architecture}
    Our approach leverages \Chronos, a pretrained TSFM designed for probabilistic
    forecasting, as a fixed feature extractor. We specifically utilize the \texttt{chronos2}
    checkpoint. The TSFM follows a decoder-only transformer architecture.

    \paragraph{Context Extraction}
    We utilize the \Chronos\ model, denoted as $\Phi_{e}$, to extract high-dimensional
    features from a windowed sensor sequence $\mathbf{X}_{t-L:t}$. By processing
    the sequence up to in a window of $L$ timesteps, we make use of \Chronos\ in-context
    learning capabilities, i.e. we extract embeddings that capture temporal
    context and dependencies across the context window of size $L$. Keeping the TSFM
    parameters frozen, we obtain a sequence of hidden representations:
    \begin{equation}
        \mathbf{H}= \Phi_{e}(\mathbf{X}_{t-L:t}) \in \mathbb{R}^{L \times h},
    \end{equation}
    where $h$ is the hidden dimension of the model and $\mathbf{H}= [\mathbf{h}_{t-L}
    , \ldots, \mathbf{h}_{t}]$.

    \paragraph{RUL Estimation}
    To predict the RUL at a specific time step $t$, we use the corresponding embedding
    vector $\mathbf{h}_{t-L:t}$ from the sequence $\mathbf{H}_{t-L:t}$ over the context
    window $L$. We employ a regressor head $g_{\phi}$ with parameters $\phi$:
    \begin{equation}
        \hat{y}_{t}= g_{\phi}(\mathbf{h}_{t-L:t}).
    \end{equation}
    The head is implemented as a Multi-Layer Perceptron (MLP) consisting of two linear
    layers with a hidden width of $m$, utilizing ReLU activation. A final ReLU is
    applied to the output to enforce $\hat{y}_{t}\ge 0$. We apply dropout with
    rate $p$ after the first hidden layer and optimize only the head parameters
    $\phi$.

    \section{Experiments}

    \subsection{Dataset and experimental setup}
    Our main experiments are conducted on industrial sensor datasets provided by
    Nokia Solutions and Networks GmbH \& Co. KG, Germany, where the failure
    events are provided in repair logs. The datasets contain multivariate sensor
    streams from two different devices; we refer to them as Device~A and Device~B.
    We preprocess the data by resampling the sensor streams to a uniform time grid,
    normalizing each sensor channel, and computing RUL labels from the repair logs
    as described in Sec.~\ref{sec:method_data}. For Device~A, after
    preprocessing we obtain a total of $297{,}345$ labeled samples. The sensors return
    multivariate sensor recordings with $D=87$, and we set the resampling step
    size to $\Delta t = 1$ hour. The context window length $L$ is set to $5$ in the
    following experiments. In addition, we evaluate on a second dataset from Device~B.
    After the same preprocessing procedure, Device~B contains $119{,}364$ annotated
    samples with $D=51$ sensor channels. We use the same resampling step size
    $\Delta t$ and context length $L$.

    To prevent temporal leakage, we employ a chronological split. The training
    set consists of windows ending at $t \le T_{\mathrm{train}}$, while the test
    set contains windows where $t > T_{\mathrm{test}}$. We select split points
    to achieve an $85{:}15$ ratio. Crucially, we discard any window
    $[t, ... , T]$ that overlaps the boundary between splits. We report performance
    on the held-out test set using Mean Absolute Error (MAE) and Mean Squared Error
    (MSE).

    \subsection{Implementation}

    We train our approach using the Mean Squared Error (MSE) loss and the Adam
    optimizer~\cite{kingma2014adam} with a learning rate of $10^{-3}$. Training runs
    for a maximum of 50 epochs with a batch size of 64. The fine-tuning process requires
    about 2 hours on a single NVIDIA A6000 GPU. We train separate models for
    device~A ($\approx300K$ parameters) and device~B ($\approx250K$ parameters).

    \subsection{Baselines}
    We compare our approach against classical, non-sequential regressors trained
    on raw data points and lightweight sequential neural baselines trained on sequences
    with context windows length $L$.

    For the non-sequential setting, we evaluate linear regression and random
    forests~\cite{hastie2009elements} on normalized sensor vectors
    $\mathbf{x}_{t}$, using the same capped target $y_{t}$ and the same train/test
    splits as our method. For sequential baselines, we train LSTM~\cite{hochreiter1997lstm}
    and GRU~\cite{gru} regressors on the windowed inputs $\mathbf{X}_{t-L:t}\in\mathbb{R}
    ^{L\times D}$ and predict $\hat{y}_{t}$ from the final recurrent state via a
    linear readout. We additionally train gradient boosting~\cite{friedman2001greedy}
    on window features $\mathbf{u}_{t}=\varphi (\mathbf{X}_{t-L:t})$, where $\varphi
    (\cdot)$ computes per-channel statistical features (mean, std, min, max,
    quantiles), last value, first differences, and a linear trend slope over the
    $L$ time steps. This baseline uses the same $L$, preprocessing pipeline, and
    chronological splits as our method.

    We further evaluate two strong convolutional and attention-based sequence models:
    a Temporal Convolutional Network (TCN)~\cite{tcn} and a Transformer encoder~\cite{transformer}
    regressor. Both models take $\mathbf{X}_{t-L:t}$ as input and output a scalar
    RUL prediction through a pooling operation over the time dimension followed by
    a linear output layer. All baseline hyperparameters are tuned on a
    validation split of $\approx 10\%$ of training data and evaluated using the
    same metrics as our method. Preprocessing statistics, such as normalization
    and clipping thresholds are computed on the training split only and applied
    unchanged to the test splits.

    \subsection{Performance on 5-step windows}
    In Table~\ref{tab:seq5_both}, we report results for sequence models trained on
    $L=5$-step windows for both Device~A and Device~B. On both datasets, moving from
    non-neural approaches (upper block) to neural network-based sequence modeling
    (lower block) yields a substantial improvement, indicating that even short
    context-windows carry significant information.
    Our {\Chronos} adaptation further improves performance across both devices:
    conditioning {\Chronos} on temporal context, extracting internal embeddings,
    and adding a dedicated regression head achieves the best overall performance
    among evaluated models.

    \begin{table}[t]
        \caption{Sequence models on 5-step windows for Device~A and Device~B.}
        \label{tab:seq5_both}
        \centering
        \begin{tabular}{lrrrr}
            \hline
                                  & \multicolumn{2}{c}{\textbf{Device~A}} & \multicolumn{2}{c}{\textbf{Device~B}} \\
            \textbf{Method}       & \textbf{MAE}$\downarrow$              & \textbf{MSE}$\downarrow$             & \textbf{MAE}$\downarrow$ & \textbf{MSE}$\downarrow$ \\
            \hline
            LinReg                & 173                                   & 44123                                & 120                      & 22123                    \\
            RandForest            & 165                                   & 48655                                & 112                      & 15655                    \\
            GrBoost               & 186                                   & 53400                                & 130                      & 33400                    \\
            \hline
            GRU                   & 91                                    & 10236                                & 102                      & 9523                     \\
            LSTM                  & 93                                    & 12400                                & 99                       & 10212                    \\
            TCN / Temporal CNN    & 88                                    & 9689                                 & 112                      & 11435                    \\
            Transformer (encoder) & 90                                    & 10500                                & 95                       & 8900                     \\
            \hline
            Ours                  & \textbf{44}                           & \textbf{6513}                        & \textbf{64}              & \textbf{7212}            \\
            \hline
        \end{tabular}
    \end{table}
    \subsection{Effect of Context Length}
    \label{sec:context_length} We study how the length of context windows
    affects \Chronos\ on both devices by varying $L$. For each $L$, we generate
    the windowed inputs $\mathbf{X}_{t-L:t}$ using the same resampling rate
    $\Delta t$ and the same chronological split protocol, keep the \Chronos\ backbone
    frozen, and retrain only the MLP head. We leave all other hyperparameters fixed.
    Each setting is evaluated after training our approach for $50$ epochs. For the
    remaining baselines, we retrain and re-tune for each $L$ using the same
    validation split.

    \begin{figure}[t]
        \centering
        \begin{tikzpicture}
            \begin{axis}[
                width=0.7\linewidth,
                height=4.5cm,
                xlabel={Context length $L$},
                ylabel={MAE $\downarrow$ },
                xmin=0,
                xmax=100,
                ymin=20,
                ymax=95,
                xtick={0,20,40,60,80,100},
                ytick distance=10,
                grid=both,
                major grid style={line width=0.25pt,draw=gray!25},
                minor grid style={line width=0.15pt,draw=gray!15},
                minor x tick num=1,
                minor y tick num=1,
                axis line style={line width=0.6pt},
                tick style={line width=0.6pt},
                tick label style={font=\footnotesize},
                label style={font=\footnotesize},
                title style={font=\footnotesize,yshift=1.0ex},
                title={Device A},
                legend style={ font=\footnotesize, at={(1.05,0.98)}, anchor=north west, row sep=-3
                pt },
                legend cell align=left,
                every axis plot/.append style={ thick, line join=round, line cap=round },
            ]
                \addplot+[ very thick, mark=*, mark size=1.8pt, ] coordinates {(0,65) (10,40) (20,36) (30,34) (40,32) (50,29) (60,26) (70,23) (80,22) (90,24) (100,28)};
                \addlegendentry{Ours}

                \addplot+[ mark=diamond*, mark size=1.6pt, dash pattern=on 6pt
                off 2pt ] coordinates
                {(0,92) (10,78) (20,70) (30,64) (40,59) (50,56) (60,54) (70,51) (80,51) (90,51) (100,52)};
                \addlegendentry{TCN}

                \addplot+[ mark=asterisk, mark size=2.2pt, dash pattern=on 1pt
                off 2pt ] coordinates
                {(0,99) (10,80) (20,72) (30,66) (40,61) (50,58) (60,53) (70,50) (80,48) (90,47) (100,48)};
                \addlegendentry{Transformer}

                \addplot+[ mark=square*, mark size=1.6pt, dash pattern=on 4pt off
                2pt ] coordinates {(0,104) (10,85) (20,80) (30,76) (40,73) (50,71) (60,69) (70,68) (80,67) (90,68) (100,69)};
                \addlegendentry{GRU}

                \addplot+[ mark=triangle*, mark size=1.7pt, dash pattern=on 2pt
                off 2pt ] coordinates
                {(0,92) (10,86) (20,82) (30,78) (40,75) (50,73) (60,72) (70,69) (80,70) (90,71) (100,72)};
                \addlegendentry{LSTM}

                \addplot+[ mark=x, mark size=2.0pt, solid ] coordinates
                {(0,92) (10,84) (20,79) (30,75) (40,72) (50,70) (60,68) (70,67) (80,66) (90,67) (100,66)};
                \addlegendentry{GradBoost}
            \end{axis}
        \end{tikzpicture}
        \caption{MAE versus context length $L$ on Device~A. Baselines show mild fluctuations;
        TCN and Transformer are the strongest baselines while Ours shows
        significant improvement over all baselines.}
        \label{fig:context_sweep_A}
    \end{figure}
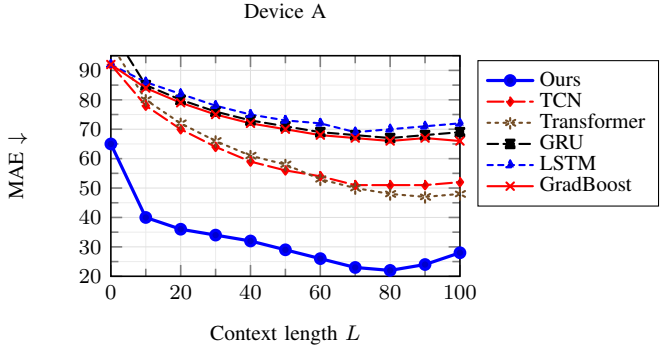
    Figure~\ref{fig:context_sweep_A} shows the effect of varying context length
    $L$ for Device~A. The performance of our approach improves significantly as $L$
    increases, with the most substantial gains observed up to $L=80$. Beyond this
    point, performance saturates. In contrast, other baselines show only mild
    fluctuations with increasing $L$, indicating that they do not effectively leverage
    longer context windows. The TCN and Transformer baselines are the strongest
    among the non-TSFM approaches, but still significantly underperform compared
    to our method across all context lengths.

    \subsection{Ablation: Regression head}
    We ablate the capacity of the supervised regression head while keeping the \Chronos\
    backbone frozen in table \ref{tab:ablation_head}. Specifically, we replace our
    default MLP head with a linear head and a deeper MLP variant (4 layers),
    using identical preprocessing, context length $L$ and training budget. All
    ablations are done on device~A.

    Our goal is to determine how much of the performance gain can be attributed
    to the pretrained \Chronos\ representations versus the supervised regression
    head. If \Chronos\ representations are the primary factor of performance,
    then a simple linear head should already achieve most of the performance
    improvement, and increasing head capacity should provide only small improvements.
    On the other hand, if performance improves as the head becomes deeper, this
    indicates that a large fraction of the mapping from inputs
    $\mathbf{X}_{t-L:t}$ to RUL $y_{t}$ is learned by the regression head rather
    than the pretrained backbone.

    When compared to the TCN baseline, even a linear head on top of frozen \Chronos\
    embeddings achieves a significant improvement (MAE $60$ vs.\ $88$). This shows
    that the pretrained backbone provides a strong representation for RUL estimation.
    The 2-layer MLP outperforms the linear head, indicating that some nonlinearity
    is beneficial for mapping \Chronos\ embeddings to RUL predictions. However, increasing
    the head capacity to 4 layers does not yield significant further improvements.

    \begin{table}[t]
        \caption{Ablation on regression head architecture.}
        \label{tab:ablation_head}
        \centering
        \begin{tabular}{lcc}
            \hline
            \textbf{Head}        & \textbf{MAE}$\downarrow$ & \textbf{MSE}$\downarrow$ \\
            \hline
            TCN (for comparison) & 88                       & 9689                     \\
            \hline
            Linear               & 60                       & 8044                     \\
            2-layer MLP          & 44                       & 6513                     \\
            4-layer MLP          & 45                       & 6342                     \\
            \hline
        \end{tabular}
    \end{table}

    \section{Conclusion}
    \label{sec:conclusion} This paper studied whether a pretrained time-series
    foundation model can serve as an effective feature extractor for RUL
    estimation in industrial predictive maintenance. In our approach, a simple
    adaptation that leverages the frozen \Chronos\ backbone , we extract context-dependent
    embeddings from multivariate sensor windows, and train a lightweight regressor
    to predict RUL. On the Nokia dataset, the approach achieved lower error than
    a broad set of classical and deep learning baselines, and an ablation over
    context length showed that longer histories can substantially improve
    performance.

    Beyond raw performance, we discussed evaluation and modeling choices that are
    critical for trustworthy RUL results. As future work, we plan to (i) leverage
    \Chronos\ probabilistic outputs to provide calibrated uncertainty for maintenance
    decisions, (ii) improve multivariate tokenization and channel handling for
    heterogeneous sensors, and (iii) evaluate robustness under missing sensors,
    operating-regime shifts, and low-label settings.

\section*{Acknowledgement}
    This work was supported by the German Federal Ministry of Education and Research (in German: Bundesministerium fur Bildung und Forschung, BMBF): project SUSTAINET-inNOvAte, 16KIS2255K.

    \bibliographystyle{IEEEtran}
    \bibliography{converted_refs}

@misc{chronos,
  title        = {Chronos: Learning the Language of Time Series},
  author       = {Ansari, A. F. and others},
  year         = {2024},
  eprint       = {2403.07815},
  archivePrefix= {arXiv},
  primaryClass = {cs.LG},
  doi          = {10.48550/arXiv.2403.07815}
}

@article{chronos2,
  title={Chronos-2: From univariate to universal forecasting},
  author={Ansari, Abdul Fatir and Shchur, Oleksandr and K{\"u}ken, Jaris and Auer, Andreas and Han, Boran and Mercado, Pedro and Rangapuram, Syama Sundar and Shen, Huibin and Stella, Lorenzo and Zhang, Xiyuan and others},
  journal={arXiv preprint arXiv:2510.15821},
  year={2025}
}

@article{shen2025visionts++,
  title={VisionTS++: Cross-Modal Time Series Foundation Model with Continual Pre-trained Vision Backbones},
  author={Shen, Lefei and Chen, Mouxiang and Liu, Xu and Fu, Han and Ren, Xiaoxue and Sun, Jianling and Li, Zhuo and Liu, Chenghao},
  journal={arXiv preprint arXiv:2508.04379},
  year={2025}
}

@article{friedman2001greedy,
  title={Greedy function approximation: a gradient boosting machine},
  author={Friedman, Jerome H},
  journal={Annals of statistics},
  pages={1189--1232},
  year={2001},
  publisher={JSTOR}
}

@inproceedings{das2024decoder,
  title={A decoder-only foundation model for time-series forecasting},
  author={Das, Abhimanyu and Kong, Weihao and Sen, Rajat and Zhou, Yichen},
  booktitle={Forty-first International Conference on Machine Learning},
  year={2024}
}

@article{nie2022time,
  title={A Time Series is Worth 64Words: Long-term Forecasting with Transformers},
  author={Nie, Y},
  journal={arXiv preprint arXiv:2211.14730},
  year={2022}
}

@article{woo2024unified,
  title={Unified training of universal time series forecasting transformers},
  author={Woo, Gerald and Liu, Chenghao and Kumar, Akshat and Xiong, Caiming and Savarese, Silvio and Sahoo, Doyen},
  year={2024},
  publisher={PMLR}
}

@article{kingma2014adam,
  title={Adam: A method for stochastic optimization},
  author={Kingma, Diederik P},
  journal={arXiv preprint arXiv:1412.6980},
  year={2014}
}

@book{hastie2009elements,
  title={The Elements of Statistical Learning: Data Mining, Inference, and Prediction},
  author={Hastie, T. and Tibshirani, R. and Friedman, J.H.},
  isbn={9780387848846},
  lccn={2008941148},
  series={Springer series in statistics},
  url={https://books.google.de/books?id=eBSgoAEACAAJ},
  year={2009},
  publisher={Springer}
}

@article{transformer,
  title={Attention is all you need},
  author={Vaswani, Ashish and Shazeer, Noam and Parmar, Niki and Uszkoreit, Jakob and Jones, Llion and Gomez, Aidan N and Kaiser, {\L}ukasz and Polosukhin, Illia},
  journal={Advances in neural information processing systems},
  volume={30},
  year={2017}
}

@inproceedings{tcn,
  title={Temporal convolutional networks: A unified approach to action segmentation},
  author={Lea, Colin and Vidal, Rene and Reiter, Austin and Hager, Gregory D},
  booktitle={European conference on computer vision},
  pages={47--54},
  year={2016},
  organization={Springer}
}

@article{gru,
  title={Empirical evaluation of gated recurrent neural networks on sequence modeling},
  author={Chung, Junyoung and Gulcehre, Caglar and Cho, KyungHyun and Bengio, Yoshua},
  journal={arXiv preprint arXiv:1412.3555},
  year={2014}
}

@article{jardine2006review,
  title   = {A review on machinery diagnostics and prognostics implementing condition-based maintenance},
  author  = {Jardine, A. K. S. and Lin, D. and Banjevic, D.},
  journal = {Mechanical Systems and Signal Processing},
  year    = {2006},
  volume  = {20},
  number  = {7},
  pages   = {1483--1510},
  doi     = {10.1016/j.ymssp.2005.09.012}
}

@article{si2011rulreview,
  title   = {Remaining useful life estimation -- A review on the statistical data driven approaches},
  author  = {Si, X.-S. and Wang, W. and Hu, C.-H. and Zhou, D.-H.},
  journal = {European Journal of Operational Research},
  year    = {2011},
  volume  = {213},
  number  = {1},
  pages   = {1--14},
  doi     = {10.1016/j.ejor.2010.11.018}
}

@inproceedings{saxena2008cmapss,
  title     = {Damage Propagation Modeling for Aircraft Engine Run-to-Failure Simulation},
  author    = {Saxena, Abhinav and Goebel, Kai and Simon, Don and Eklund, Neil},
  booktitle = {2008 International Conference on Prognostics and Health Management},
  year      = {2008},
  address   = {Denver, CO, USA},
  publisher = {IEEE},
  pages     = {1--9},
  doi       = {10.1109/PHM.2008.4711414}
}

@article{hochreiter1997lstm,
  title   = {Long Short-Term Memory},
  author  = {Hochreiter, Sepp and Schmidhuber, J{\"u}rgen},
  journal = {Neural Computation},
  year    = {1997},
  volume  = {9},
  number  = {8},
  pages   = {1735--1780},
  doi     = {10.1162/neco.1997.9.8.1735}
}

@inproceedings{cho2014gru,
  title     = {Learning Phrase Representations using {RNN} Encoder--Decoder for Statistical Machine Translation},
  author    = {Cho, Kyunghyun and van Merri{\"e}nboer, Bart and Gulcehre, Caglar and Bahdanau, Dzmitry and Bougares, Fethi and Schwenk, Holger and Bengio, Yoshua},
  booktitle = {Proceedings of the 2014 Conference on Empirical Methods in Natural Language Processing (EMNLP)},
  year      = {2014},
  address   = {Doha, Qatar},
  publisher = {Association for Computational Linguistics},
  pages     = {1724--1734},
  doi       = {10.3115/v1/D14-1179},
  url       = {https://aclanthology.org/D14-1179/}
}

@article{li2018cnn_rul,
  title   = {Remaining useful life estimation in prognostics using deep convolution neural networks},
  author  = {Li, Xiang and Ding, Qian and Sun, Jian-Qiao},
  journal = {Reliability Engineering \& System Safety},
  year    = {2018},
  volume  = {172},
  pages   = {1--11},
  doi     = {10.1016/j.ress.2017.11.021}
}

@INPROCEEDINGS{wiederer_benchmark,
  author={Wiederer, Julian and Schmidt, Julian and Kressel, Ulrich and Dietmayer, Klaus and Belagiannis, Vasileios},
  booktitle={2022 IEEE 25th International Conference on Intelligent Transportation Systems (ITSC)}, 
  title={A Benchmark for Unsupervised Anomaly Detection in Multi-Agent Trajectories}, 
  year={2022},
  volume={},
  number={},
  pages={130-137},
  doi={10.1109/ITSC55140.2022.9922440}}

@INPROCEEDINGS{casas_adversarial,
  author={Casas, Leslie and Klimmek, Attila and Navab, Nassir and Belagiannis, Vasileios},
  booktitle={2020 28th European Signal Processing Conference (EUSIPCO)}, 
  title={Adversarial Signal Denoising with Encoder-Decoder Networks}, 
  year={2021},
  volume={},
  number={},
  pages={1467-1471},
  doi={10.23919/Eusipco47968.2020.9287738}}

@article{fan2024star,
  title   = {A Two-Stage Attention-Based Hierarchical Transformer for Turbofan Engine Remaining Useful Life Prediction},
  author  = {Fan, Zhengyang and Li, Wanru and Chang, Kuo-Chu},
  journal = {Sensors},
  year    = {2024},
  volume  = {24},
  number  = {3},
  pages   = {824},
  doi     = {10.3390/s24030824}
}

@article{fu2024dualmixer,
  title         = {Supervised Contrastive Learning based Dual-Mixer Model for Remaining Useful Life Prediction},
  author        = {Fu, En and others},
  year          = {2024},
  eprint        = {2401.16462},
  archivePrefix = {arXiv},
  primaryClass  = {cs.LG},
  url           = {https://arxiv.org/abs/2401.16462}
}

@article{du2024madan,
  title   = {Remaining useful life prediction under variable operating conditions via multisource adversarial domain adaptation networks},
  author  = {Du, Junrong and Song, Lei and Gui, Xuanang and Zhang, Jian and Guo, Lili and Li, Xuzhi},
  journal = {Applied Soft Computing},
  year    = {2024},
  doi     = {10.1016/j.asoc.2024.111717}
}

@article{jiang2024stahpinn,
  title         = {Spatio-temporal Attention-based Hidden Physics-informed Neural Network for Remaining Useful Life Prediction},
  author        = {Jiang, Feilong and Hou, Xiaonan and Xia, Min},
  year          = {2024},
  eprint        = {2405.12377},
  archivePrefix = {arXiv},
  primaryClass  = {cs.LG},
  url           = {https://arxiv.org/abs/2405.12377}
}

@article{wang2024diffrul,
  title   = {Data augmentation based on diffusion probabilistic model for remaining useful life estimation of aero-engines},
  author  = {Wang, Wei and Song, Honghao and Si, Shubin and Lu, Wenhao and Cai, Zhiqiang},
  journal = {Reliability Engineering \& System Safety},
  year    = {2024},
  volume  = {252},
  doi     = {10.1016/j.ress.2024.110394}
}

@article{wen2025diffusion,
  title   = {A generalized diffusion model for remaining useful life prediction with uncertainty},
  author  = {Wen, Bincheng and Zhao, Xin and Tang, Xilang and Xiao, Mingqing and Zhu, Haizhen and Li, Jianfeng},
  journal = {Complex \& Intelligent Systems},
  year    = {2025},
  doi     = {10.1007/s40747-024-01773-w}
}

@article{feng2025kairos,
  title={Kairos: Towards Adaptive and Generalizable Time Series Foundation Models},
  author={Feng, Kun and Lan, Shaocheng and Fang, Yuchen and He, Wenchao and Ma, Lintao and Lu, Xingyu and Ren, Kan},
  journal={arXiv preprint arXiv:2509.25826},
  year={2025}
}
\end{document}